\documentclass[runningheads]{llncs}
\usepackage{graphicx}
\usepackage{amsmath,amssymb}
\usepackage{comment}
\usepackage[colorinlistoftodos]{todonotes}
\usepackage{caption}
\usepackage{subcaption}
\usepackage{xcolor}
\usepackage[columnwise]{lineno}
\usepackage{textcmds}

\usepackage{geometry}
\begin{document}

\title{Regenerating Soft Robots through\\Neural Cellular Automata}
\author{Kazuya Horibe\inst{1,2,3}, Kathryn Walker\inst{1}, \and Sebastian Risi\inst{1}}
\authorrunning{K. Horibe et al.}
 \institute{IT University of Copenhagen, Copenhagen, Denmark \and Osaka University, Osaka, Japan \and Cross Labs, Cross Compass Ltd., Tokyo, Japan}
\maketitle              % typeset the header of the contribution
\begin{abstract}

%The abstract should briefly summarize the contents of the paper in 15--250 words.
Morphological regeneration is an important feature that highlights the environmental adaptive capacity of biological systems. Lack of this regenerative capacity significantly limits the resilience of machines and the environments they can operate in. To aid in addressing this gap, we develop an approach for simulated soft robots to regrow parts of their morphology when being damaged. Although numerical simulations using soft robots have played an important role in their design, evolving soft robots with regenerative capabilities have so far received comparable little attention. Here we propose a model for soft robots that regenerate through a neural cellular automata. Importantly, this approach only relies on local cell information to regrow damaged components, opening interesting possibilities for physical regenerable soft robots in the future. Our approach allows simulated soft robots that are damaged to partially regenerate their original morphology through local cell interactions alone and regain some of their ability to locomote. These results take a step towards equipping artificial systems with regenerative capacities and could potentially allow for more robust operations in a variety of situations and environments. The code for the experiments in this paper is available at: \url{github.com/KazuyaHoribe/RegeneratingSoftRobots}.

\end{abstract}
\section{Introduction}
%Regeneration in biological systems
Many organisms have regenerative capabilities, allowing them to repair and reconfigure their morphology in response to damage or changes in components \cite{Carlson2011}. For example, the primitive organisms Hydra and Planaria are particularly capable of regeneration and can thus achieve complete repair, no matter what location of the body part is cut off \cite{Vogg2019,Levin2019}. Furthermore, salamanders are capable of regenerating an amputated leg \cite{Vieira2020}. Many biological systems achieve regeneration by retaining information on the damaged parts \cite{Blackiston2015}.

While biological systems are surprisingly robust, current robotic systems are fragile and often not able to recover from even minor damage. 
Furthermore, the majority of damage recovery approaches in robotics has focused on damage compensation through behavioral changes alone~\cite{Chatzilygeroudis2018,Cully2015,Kano2017,Ren2015,Kwiatkowski2019a}; damage recovery through the regrowth of morphology has received comparable little attention.

% Soft robots for regeneration
In this present study, we develop a neural cellular automata approach for soft robot locomotion and morphological regeneration. 
Cellular automata (CA) were first proposed by Neumann and Ulam in the 1940s 
and consist of a regular grid of cells where each cell can be in any one of a finite set of states \cite{Neumann1966}. Each cell determines its next state based on local information (i.e.\ the states of its neighboring cells) according to pre-defined rules.
In a neural cellular automata, instead of having hand-designed rules, a neural network learns the update rules~\cite{Cenek2013,Miller2004b,Wulff1992}.    
In a recent impressive demonstration of a neural CA,  Mordvintsev et al. trained a neural CA to grow complex two-dimensional images starting from a few initial cells \cite{Mordvintsev2020}. In addition, the authors successfully trained the system  to recover the pattern, when parts of it were removed (i.e.\ it was able to regrow the target pattern). The neural network in their work is a convolutional network, which lends itself to represent neural CAs \cite{Gilpin2019}. 
Earlier work by Miller showed that automatically recovery of simpler damaged target patterns is also possible with genetic programming \cite{Miller2004b}. 

% Explain our model
In this study, we extend the neural CA approach to simulated soft robots, which develop from a single cell, and are able to evolve the ability to locomote and regenerate. The results show that when the simulated soft robots are partially damaged, they are capable to move again by regrowing a morphology close to their original one. 
Our approach opens up interesting future research directions for more resilient soft robots that could ultimately be transferred to the real world. 

\section{Related work}
\subsection{Evolved virtual creatures} 

The evolution of virtual creatures first began with Karl Sim's seminal work nearly three decades ago~\cite{Sims1994c}, with creatures composed of blocks interacting with their environment and other individuals in a virtual physical space, evolving their own body plans. 
Since then many researchers have explored the use of artificial evolution to train virtual creatures and even transferred some of these designs to the real world~\cite{Dellaert2003,Eggenberger1997,Ostergaard2003,Hornby2001,Lipson2000,Radhakrishna2018,Risi2013}. It should be noted, however, that in each of the above examples, the morphology of the evolved robot is fixed; it does not develop over its lifetime. 

More recently, this research field has embraced  approaches based on compositional pattern producing networks (CPPN) \cite{Stanley2007b,Cheney2013a,Cellucci2017}, which are a special kind of neural network.  Furthermore, using  CPPN-based approaches, researchers have been able to explore evo-devo virtual creatures, where development continues during interaction with the environment, further increasing the complexity of the final body plans~\cite{Kriegman2018,Kriegman2017}. However, the lifetime development of these creature tends to be limited to material properties, rather than growth of complete body parts.

Kriegman et al. also proposed a modular soft robot automated design and construction framework~\cite{Kriegman2020b}. The framework's ability to transfer robot designs from simulation to reality could be a good match for our neural CA method in the future, which increases morphological complexity  during development.

\subsection{Cellular automata}
Instead of a CPPN-based approach, which relies on having access to a global coordinate system, we employ a neural cellular automata to grow virtual soft robots  solely based on the local interaction of cells.  As previously discussed, cellular automata (CA) were first studied by Neumann and Ulam \cite{Neumann1966} in the 1940s, taking inspiration from observations of living organisms. When correctly designed, CAs have been able to reproduce some of the patterns of  growth, self-replication, and self-repair of natural organisms, only through local cell interactions.

Wolfram exhaustively examined the rules of one-dimensional CAs and classified them according to their behaviors~\cite{Wolfram1983}. Later, Langton discovered that behavior of CAs could be determined with a single parameter~\cite{Langton1990}.  Similar dominant parameters and behaviors have been searched for in two-dimensional CAs  using tools from information theory and dynamical systems theory ~\cite{Packard1985}. 

More recently, optimization methods (e.g.\ evolutionary algorithms, gradient descent) have been employed to train neural networks that in turn dictate the behaviour (i.e.\ growth rules) of a CA. Such a CA is called a neural cellular automata \cite{Mordvintsev2020,Nichele2018,Wulff1992}. Neural CAs are able to learn complex rules, which enable growth to difficult 2D target patterns~\cite{Miller2004b,Nichele2018}, and can also regrow patterns when they are  partially removed~\cite{Mordvintsev2020}.  In this paper, we extend the work on neural CAs to soft robots, which can move and regrow once their morphologies are damaged. 

\subsection{Shape changing robots and damage recovery}
%There have been numerous examples of robots that are able to mimic the morphological development of organisms that we see in nature and similar to what we present here.
Krigeman et al. evolved robots that were capable of adapting their resting volume of each voxel in response to environmental stress~\cite{Kriegman2019a}. In recent years, not only the locomotion performance of organisms, but also their environmental adaptability through shape change has attracted attention. For a recent review of approaches that allow robots to transform in order to cope with different shapes and tasks, see Shah et al.~\cite{Shah2020}.

In terms of damage recovery, traditionally approaches have focused on the robot's control system to combat loss of performance. Building on ideas from morphological computation and embodiment, more recently morphological change has been investigated as a mechanism for damage recovery.
Such work includes that by Kriegman et al.\cite{Kriegman2019a}, where silicone based physical voxel robots were able to recover from voxel removal. Furthermore, Xenobots, synthetic creatures designed from biological tissue \cite{Kriegman2020b}, have shown to be capable of reattachment (i.e.\  healing after insult). 

Our model uses only local information (i.e.\ each cell only communicates with its neighbors) and could be applied as a design method for regenerating soft robots composed of biological tissue using techniques which control gene expression and bioelectric signaling \cite{Blackiston2013,Thompson2014}. 
We believe that by using only local information, our method is particularly biologically plausible and therefore might work on real robots in the future, with the help of various biological tissues editing technologies. In particular, an exciting direction is to combine the approach with biological robots such as the Xenobots \cite{Kriegman2020b}.

\section{Growing Soft Robots with Neural Cellular Automata}
\begin{figure}[h]
    \begin{center}
        \includegraphics[width=4in,angle=0]{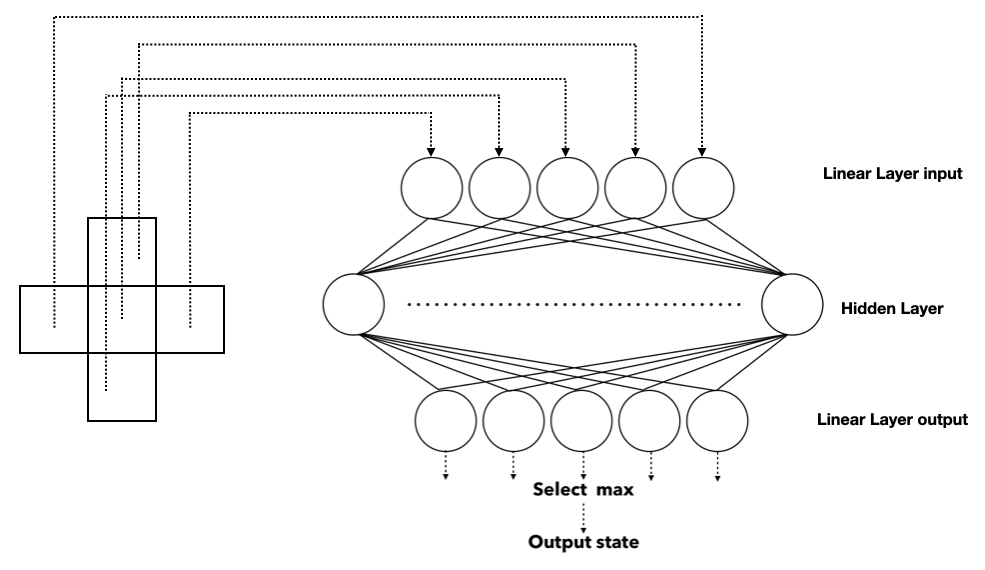}
        \caption{\textbf{Neural CA representation.} The center cell and its neighboring cells are shown.  Each cell state is an input to the neural network (bottom).  The center cell transitions to the state with the highest network output.}
        \label{Neural Cellular Automata}
    \end{center}
\end{figure}

The neural CA representation for our soft robot is shown in Fig.~\ref{Neural Cellular Automata}. For each cell, the same network maps the neighborhood cell's input to a new cell state.  The cell states are discrete values from a finite set, which we map to a continuous value before passing it to the neural network. The dimension of the input layer corresponds to the number of cell neighbors (e.g.\ Neumann and Moor neighborhood). The neural network has one output for each possible cell state and is assigned the state that corresponds to the largest activated output.
%And then repeat the next step state determination by this neural network for all cells\todo{last sentence is unclear}.

We use a  three-layer networks with tanh activation functions. %\todo{how many neurons per layer?} 
We experiment with both \textbf{feed forward} (the hidden layer is a linear layer) and \textbf{recurrent networks} (the hidden layer is an LSTM unit~\cite{Hochreiter1997}), which means that each cell has its own memory. 
The dimension of the hidden layer in this paper is set to $64$ unless otherwise noted.
The recurrent setup is inspired by recent experimental reports that organisms store information about the original morphology in a distributed manner in the bioelectrical signaling networks~\cite{Levin2017,McLaughlin2018}.

Following Mordvintsev et al.~\cite{Mordvintsev2020}, the network has an additional alpha channel output ($\alpha$)  that determines the maturity of a cell. If $\alpha$ is greater than 0.1, the cell is tagged as ``living''. A cell with $0 < \alpha < 0.1$ is tagged as ``growing''. When the neural network is calculating the next state of a cell, the inputs from empty or growing cells are set to 0.0. This way, it is possible to gradually expand the area of mature cells from a few initial cells. Mature cells can also die and become empty; to guarantee that robots  do not have any isolated and unconnected cells, isolated cells are removed before the robot is evaluated following the procedure below. First, we remove cells that are connected diagonally but not horizontally to eliminate physical simulation error in VoxCad. Second, we remove independent cells which do not have any neighboring cells. Both of these two steps are applied to each cell sequentially in the order of their indexes. %\todo{how do we calculate if a cell is isolated?}

\subsection{Soft Robot Simulator}
All soft robot experiments are performed in the open-source physical simulator Voxelyze~\cite{Hiller2009}.
We consider a locomotion task for soft robots composed of a $7 \times 7$ and $9 \times 9 \times 9$ grid of voxels in 2D and 3D, respectively. 
We adopt the Dirichlet boundary condition, and the cell state of the outer frame is always empty. Thus, the actual maximum size of the soft robots is $5\times5$ (Fig.~\ref{2d creature development}) and $7\times7\times7$ (Fig.~\ref{3d creature development}). 
At any given time, a robot is completely specified by an array of resting volumes, one for each of its $7\times7=49$ and $9\times9\times9=729$ voxels. Each voxel is volumetrically actuated according to a global signal that varies sinusoidally in volume over time. 
The actuation is a linear expansion and contraction from their resting volume.

Following Cheney et al.~\cite{Cheney2013a}, there are four types of voxels, denoted by their color: Red colored voxels can be thought of as muscle; they actively contract and expand sinusoidally at a constant frequency. Green colored voxels can also be though of as muscle, with their activation in counter-phase to red voxels. Dark blue colored voxels can be considered as bone; they are unable to expand and contract like the ``muscle" voxels. Furthermore, they have a high stiffness value. The final type of voxel used in this experiment are colored light blue voxels. These are also passive but have a lower stiffness than dark blue bone voxels. The physical and environmental Voxelyze parameters also follow the settings in Cheney et al.~\cite{Cheney2013a}.

Soft robots are evaluated for their ability to locomote for $0.25$ seconds, or $10$ actuation cycles in 2D (Fig.~\ref{2d creature locomotion}) and for $0.5$ seconds, or $20$ actuation cycles in 3D (Fig.~\ref{2D Phylum}, \ref{3D Phylum}). The evaluation times try to strike a balance between reducing computational costs while still giving sufficient time to observe interesting locomotion behaviours. Fitness is determined as the distance the robot’s center of mass moves in $0.25$ or $0.5$ seconds. 
The distance is measured in units that correspond to the length of a voxel with volume one.   
Creatures with zero voxels after their growth are automatically assigned a fitness of 0.0.

\subsection{Genetic algorithm}
To evolve a neural CA, we use a simple genetic algorithm~\cite{Holland1992,Eiben2003} that can train deep neural networks~\cite{Such2017}. The implemented GA variant performs truncation selection with the top $T$ individuals becoming the parents of the next generation. The following procedure is repeated at each generation: First, parents are selected uniformly at random. They are mutated by adding Gaussian noise to the weight vector of the neural network (its genotype): $\theta' = \theta + \sigma \epsilon$, where $\epsilon$ is drawn from $N(0, I)$ and  $\sigma $ is set to $0.03$. Following a technique called elitism,  top $N^{th}$ individuals are passed on to the next generation without mutation. 

\section{Results}

\subsection{Evolving 2D soft robots}

\begin{figure}[htpb!]
    \centering
   
    \begin{minipage}{.45\linewidth}
        \begin{subfigure}[t]{.9\linewidth}
            \includegraphics[width=\textwidth]{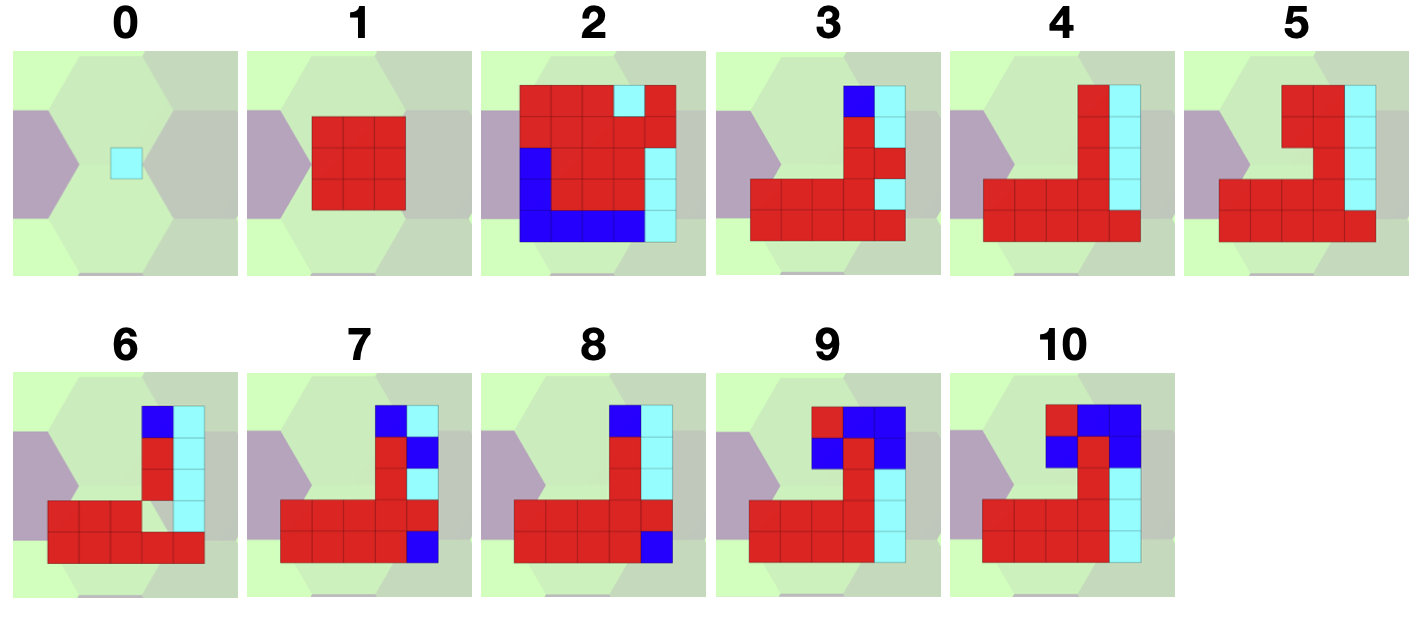}
            \caption{$2$D robot development}
            \label{2d creature development}
        \end{subfigure} \\
        \begin{subfigure}[b]{.9\linewidth}
            \includegraphics[width=\textwidth]{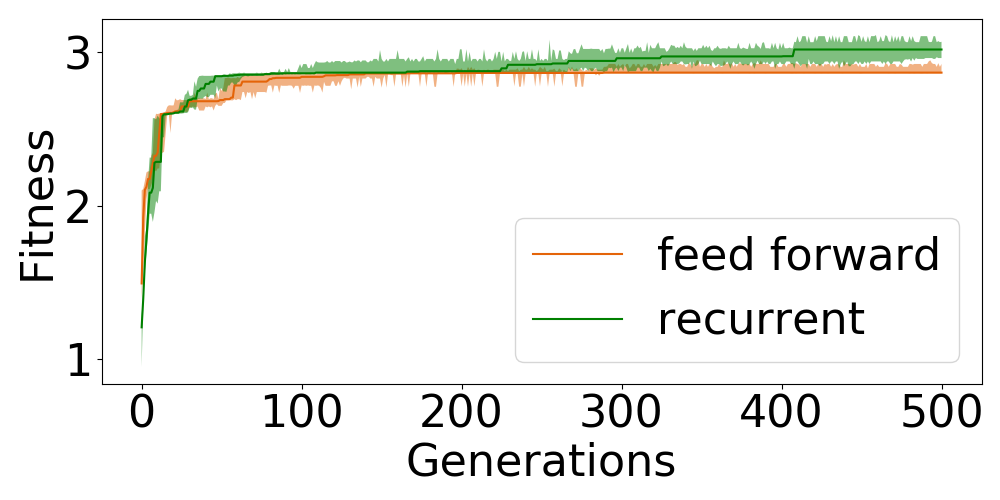}
             \caption{Training}
            \label{fitness2d}
        \end{subfigure} 
    \end{minipage}
     \begin{minipage}{.45\linewidth}
            \begin{subfigure}[t]{.9\linewidth}
                \includegraphics[width=\textwidth]{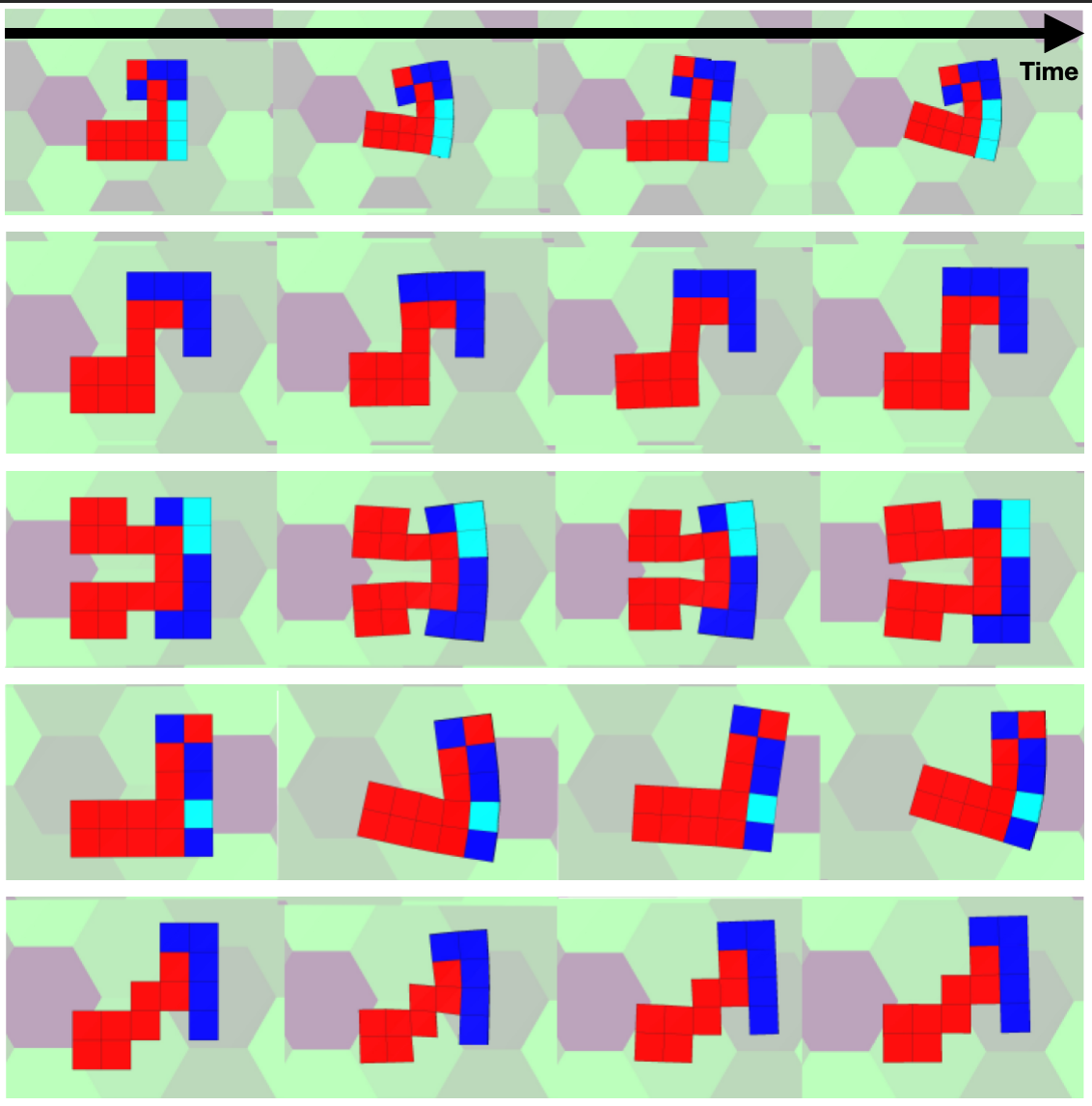}
                \caption{$2$D  robot locomotion}
                \label{2d creature locomotion}
            \end{subfigure}
    \end{minipage}
    \caption{{\bf Evolution of 2D soft robots} (a) Development of 2D soft robots through a neural cellular automata. (b) Training fitness for the  recurrent/feed forward setup. (c) Time series of soft robot behaviors as they move from left to right. From top to bottom, we refer to them as Hook type, S-type, Biped, L-type, and Zigzag.}
    \label{2d results}
\end{figure}

%In this section we present the results from our experiments in 2D.
To confirm the promise of neural CAs for growing soft robots, we first apply them to simpler 2D robot variants. Here, robots have a maximum size of $7\times7$ voxels. Since the neural CA used a Moor neighborhood, the input dimension of the neural network was $9\times2=10$, which includes neighboring cell types and alpha values.  
The output layer has a size of 6, 5 for the different states of the cell (empty=0, light blue=1, dark blue=2, red=3, green=4) plus one alpha channel. The first single soft $\&$ passive cell is placed at position $(3,3)$ and $10$ steps of development are performed. As result, 11 morphologies are obtained. (Fig.~\ref{2d creature development}). Afterwards, the final grown robot is tested in the physical simulator and allowed to attempt locomotion for 0.25 seconds (10 actuation cycles). The fitness of each robot is taken to be distance travelled by the robot from its starting point. These 2D experiments use a population size of $300$, running for 500 generations. One evolutionary run on 8 CPUs took around 12 hours.

Results are shown in Fig.~\ref{2d results}, which were obtained from ten independent evolutionary runs, using both recurrent and feed forward networks. The training mean together with bootstrapped $95\%$ confidence intervals is shown in Fig.~\ref{fitness2d}. 

Evolution produced a variety of soft robots (Fig.~\ref{2d creature locomotion}). 
A  \qq{Hook} type is distinguished by its hook-like form and locomotion, which shakes the two sides of the hook and proceed to hook the remaining one side to the floor. 
The \qq{S} shaped-robot is distinguished by its sharp and peristaltic motion with amplitude in the same direction as the direction of travel.
The \qq{Biped} has two legs and its locomotion resembles that of a frog, with the two legs pushing the robot forward.  The \qq{L} type displays a sharp and winged movement. Finally, the \qq{Zig-zag} shows a spring-like movement by stretching and retracting the zigzag structure. Enabling the cells to keep a memory of recent developmental states through a recurrent network improved performance, although only slightly (Fig.~\ref{fitness2d}). Investigating what information the evolved LSTM-based network is keeping track of during development is an interesting future research direction. 

%The reason for this result could be attributed to the balance between two factors that escape the local solution: task complexity and memory effectiveness. Task complexity can be thought of as the number of local solutions in the fitness landscape of the evolutionary calculation. On the other hand, the effectiveness of the memory could work better with a smaller number of these local solutions. It can be inferred that in the present 2D soft robot, this balance was struck by a higher degree of adaptability of the recurrent condition because the memory effectiveness prevailed.%\todo{not clear} 

\subsection{Evolving 3D soft robots}

\begin{figure}[t]%[htpb!]
    \begin{subfigure}{.5\textwidth}
        \centering
        % include first image
        \includegraphics[width=.8\linewidth]{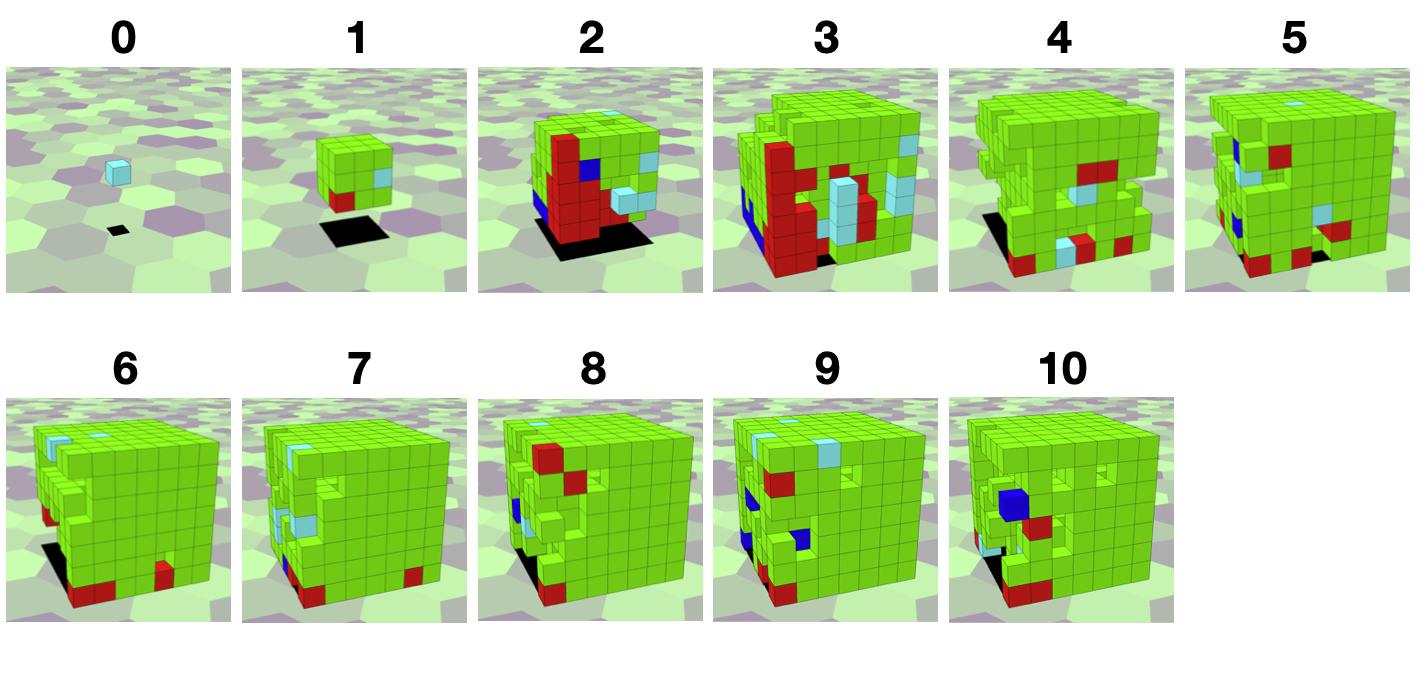}  
        \caption{3D soft robot  development}
        \label{3d creature development}
    \end{subfigure}
    \begin{subfigure}{.5\textwidth}
        \centering
        % include first image
        \includegraphics[width=.9\linewidth]{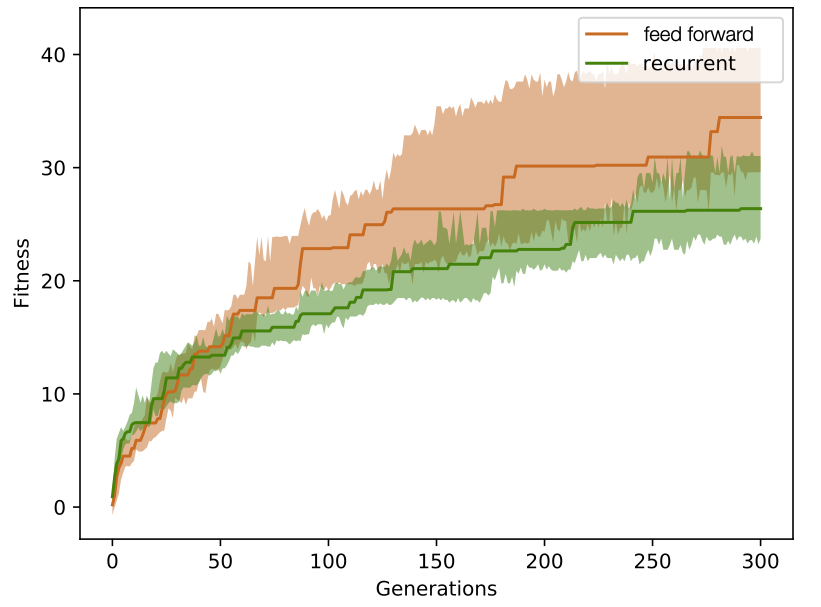}  
        \caption{Training}
        \label{fitness3d}
    \end{subfigure}
    \newline
    \begin{subfigure}{.5\textwidth}
        \centering
        % include second image
        \includegraphics[width=.8\linewidth]{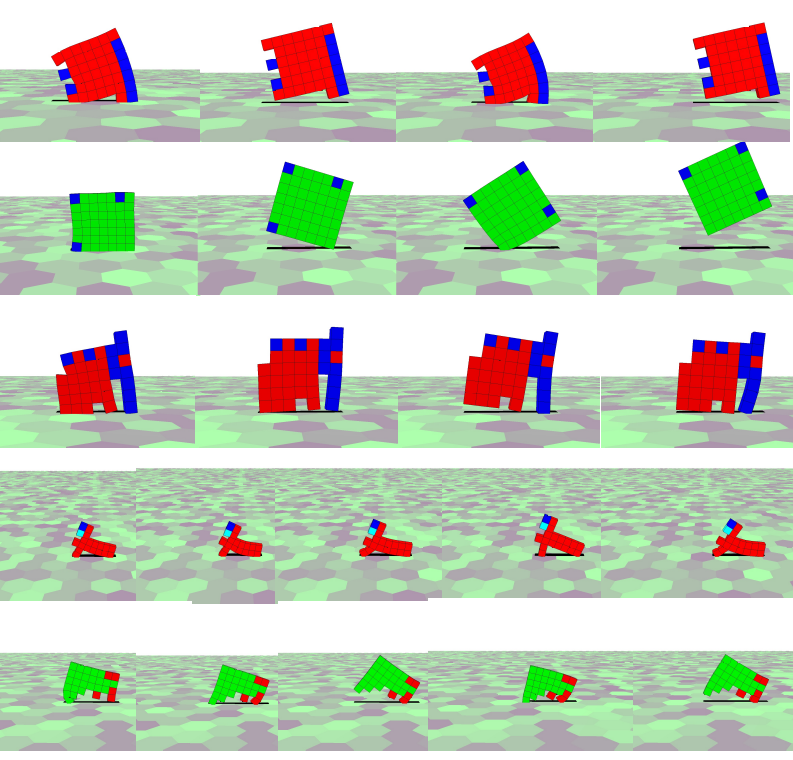}
        \caption{2D Group}
        \label{2D Phylum}
    \end{subfigure}
    \begin{subfigure}{.5\textwidth}
        \centering
        % include second image
        \includegraphics[width=.8\linewidth]{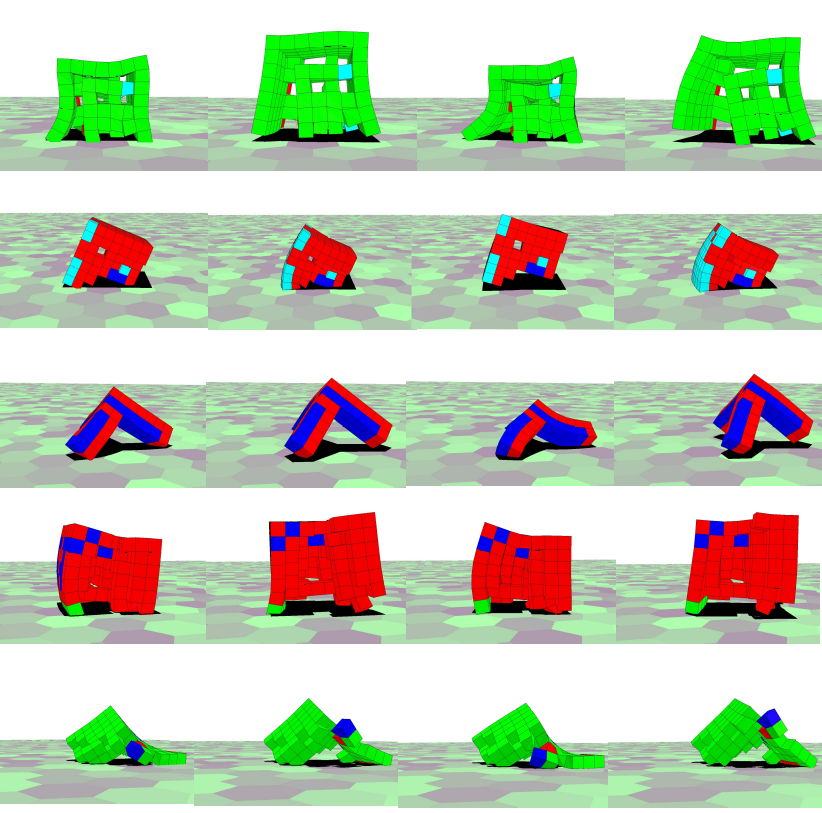}
        \caption{3D Group}
        \label{3D Phylum}   
    \end{subfigure}
    \caption{{\bf Evolution of soft robots} (a) A robots shown at different timesteps during its development. (b) Fitness over generations for the  recurrent/feed forward setup. (c) Time series of common 2D soft robot behaviors as they move from left to right. From top to bottom, we refer to them as Jumper, Roller, Pull-Push, Slider, and Jitter. (d) Common grown 3D robots: Pull-Push, L-Walker, Jumper, Crawler, and Slider.}
    \label{3d results}
\end{figure}

In this section we now extend our methodology to grow 3D robots. For these 3D robots the maximum size of the morphologies is  $9\times9\times9$. Since the neural CA uses a Moor neighborhood, the input dimension of the neural network is $3\times9\times2=54$. The hidden layer is set to $64$. The output layer is set to $5+1=6$ dimensions with the number of states of the cell and the value of its own next step alpha value. 
The first single soft $\&$ passive cell is placed at position $(4,4,4)$ and $10$ steps of growth are performed 
%; 11 morphologies are obtained. 
(Fig.~\ref{3d creature development}). The final soft robot grown after $10$ steps is tested in the physical simulated and, as with the 2D robots, the distance of the robot's center of gravity from its starting point was used as part of the fitness function. Additionally we include a voxel cost in the fitness calculation: $Fitness = (Distance) - (Voxels Cost)$. We added a ``voxel'' cost because preliminary results indicated that without this additional metric all the soft robots simply acquired a box-like morphology. Including the voxel cost metric increased diversity in the population. Note that voxel cost is the number of voxels that are neither empty nor dead.

For our 3D experiments the evaluation time is increased to $0.5$s for $20$ actuation cycles to adjust for the increased complexity of the robots. Each generation has a population size of $100$ and the next generation is  selected from the top $20\%$. The number of generations is set to 300. Note that both the generation number and population size are reduced from those values used in the 2D experiments as simulated the larger 3D robots has a higher computational cost. One evolutionary run on 1 CPU took around 80--90 hours.

Results are based on $24$ independent evolutionary runs for both the  recurrent and feed forward treatment (Fig.~\ref{fitness3d}). 
Interestingly, the feed forward setup for the 3D robots has a higher fitness than the recurrent one, in contrast to the 2D soft robot results (Fig.~\ref{fitness2d}). We hypothesize that with the increased numbers of neighbors in 3D and more complex patterns, it might be harder to evolve an LSTM-based network that can use its memory component effectively. Because the dynamics of LSTM-based networks are inherently difficult to analyse, more experiments are needed to investigate this discrepancy further.

Similarly to CPPN-encoded soft robots \cite{Cheney2013a}, 3D robots grown by an evolved neural CA (Figure~\ref{3d results}) can be classified into two groups: the first group is the two-dimensional group of organisms (Fig.~\ref{2D Phylum}), where planar morphology was acquired by evolution. Exemplary classes of locomotion in this group include the jumper, which is often composed of a single type of muscle voxel. Once a soft robot sinks down, it use this recoil to bounce up into the air and move forward. The morphology determines the angle of bounce and fall.
The Roller is similar to a square; it moves in one direction by rotating and jumping around the corners of the square. The Push-Pull is a widely seen locomotion style. A soft robot pushes itself forward with its hind legs. During this push, it pulls itself forward, usually by hooking its front legs on the ground.
The Slider has a front foot and a hind foot, and by opening and closing the two feet, it slides forward across the floor. The two legs are usually made of a single material. The Jitter moves by bouncing up and down from its hind legs to back. It has an elongated form and is often composed of a single type of muscle voxel.
The second group is the three-dimensional group of organisms, as shown in Fig.~\ref{3D Phylum}. The L-Walker resembles an L-shaped form; it moves by opening and closing the front and rear legs connected to its pivot point at the bend of the L.
The Crawler has multiple short legs and its legs move forward in concert.

\subsection{Regenerating soft robots}
\begin{figure}[htpb!]
    \begin{subfigure}{.5\textwidth}
        \centering
        \includegraphics[width=.8\linewidth]{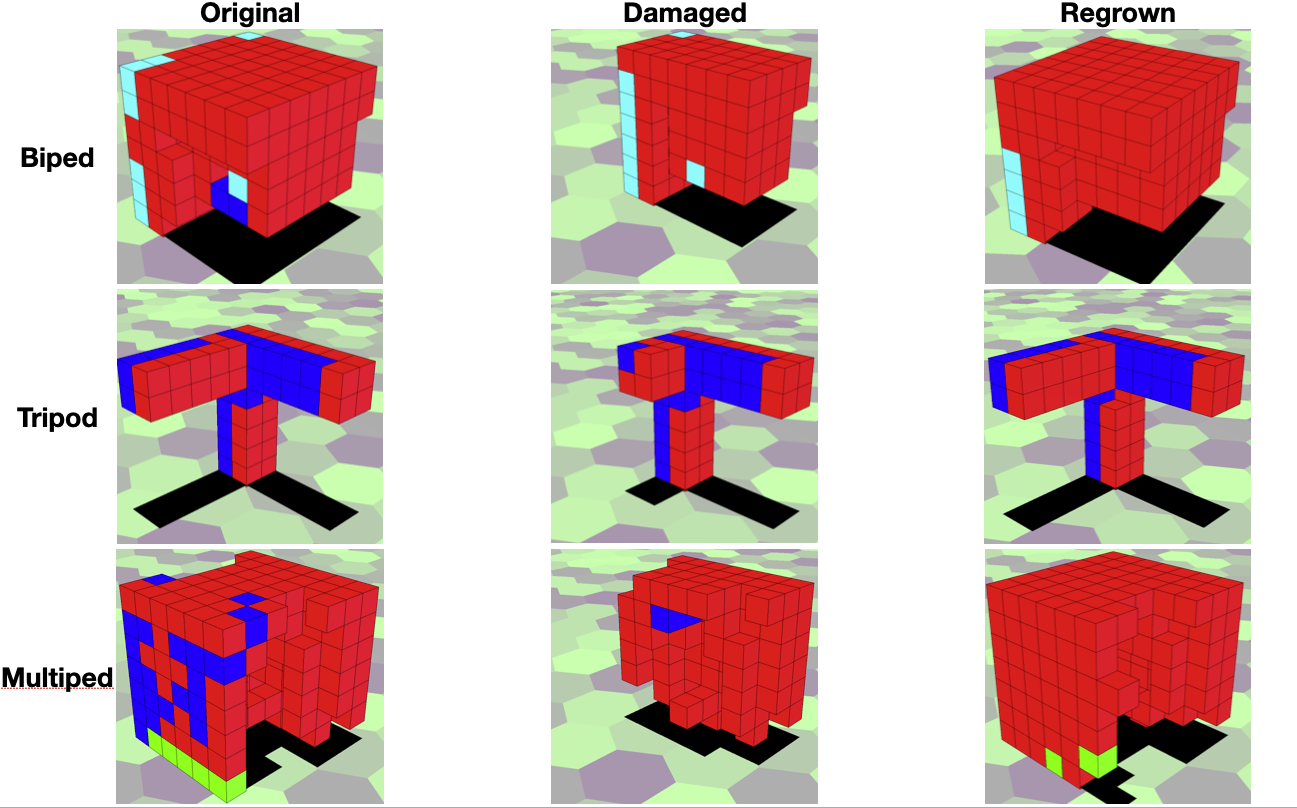}
        \caption{Original, damaged and regrown}
        \label{regeneration set up}
    \end{subfigure}
      \begin{subfigure}{.5\textwidth}
        \centering
        \includegraphics[width=1\linewidth]{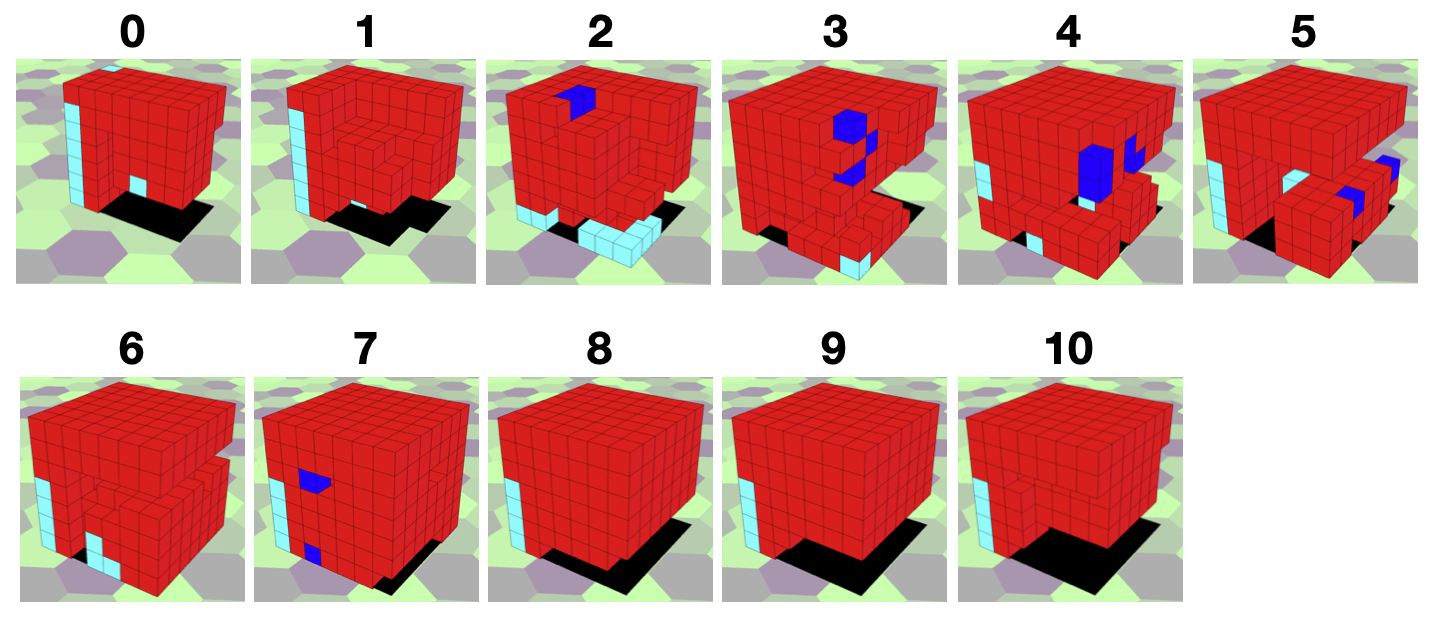}
        \caption{Biped regereration}
        \label{biped regeneration}
    \end{subfigure}
    \newline
    \begin{subfigure}{.5\textwidth}
        \centering
        \includegraphics[width=1\linewidth]{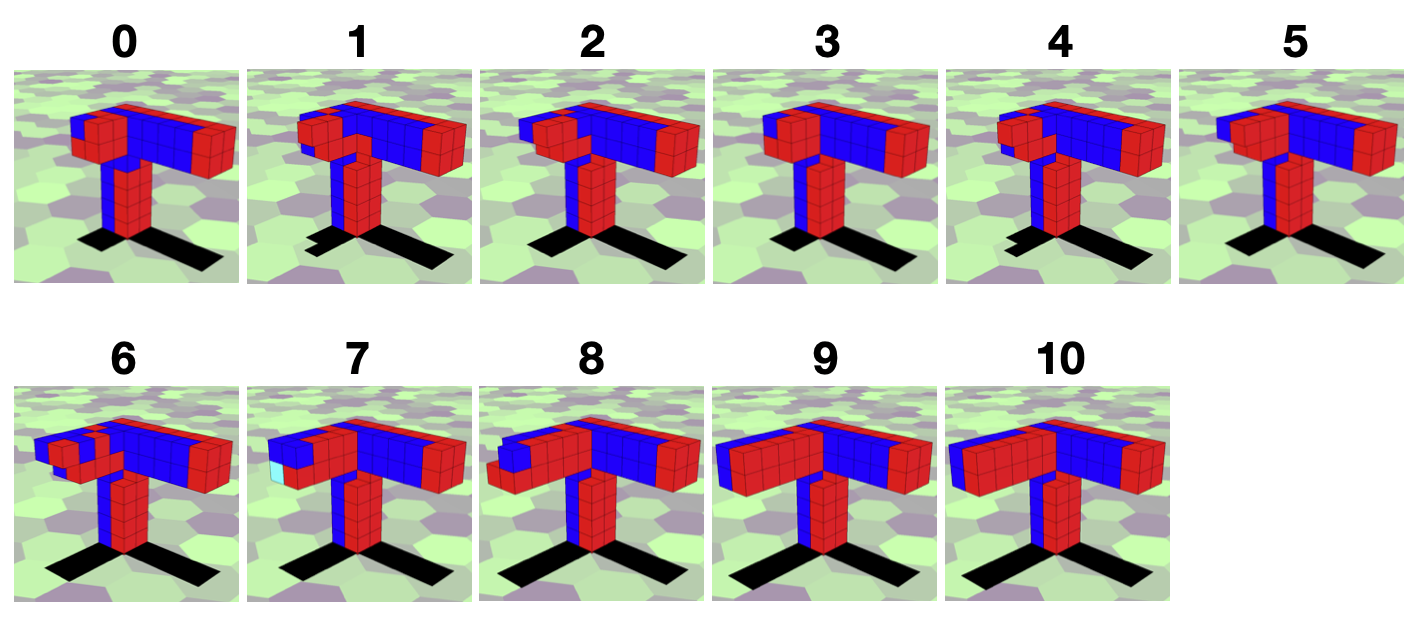}
        \caption{Tripod regereration}
        \label{Tripod regeneration}
    \end{subfigure}
    \begin{subfigure}{.5\textwidth}
        \centering
        \includegraphics[width=1\linewidth]{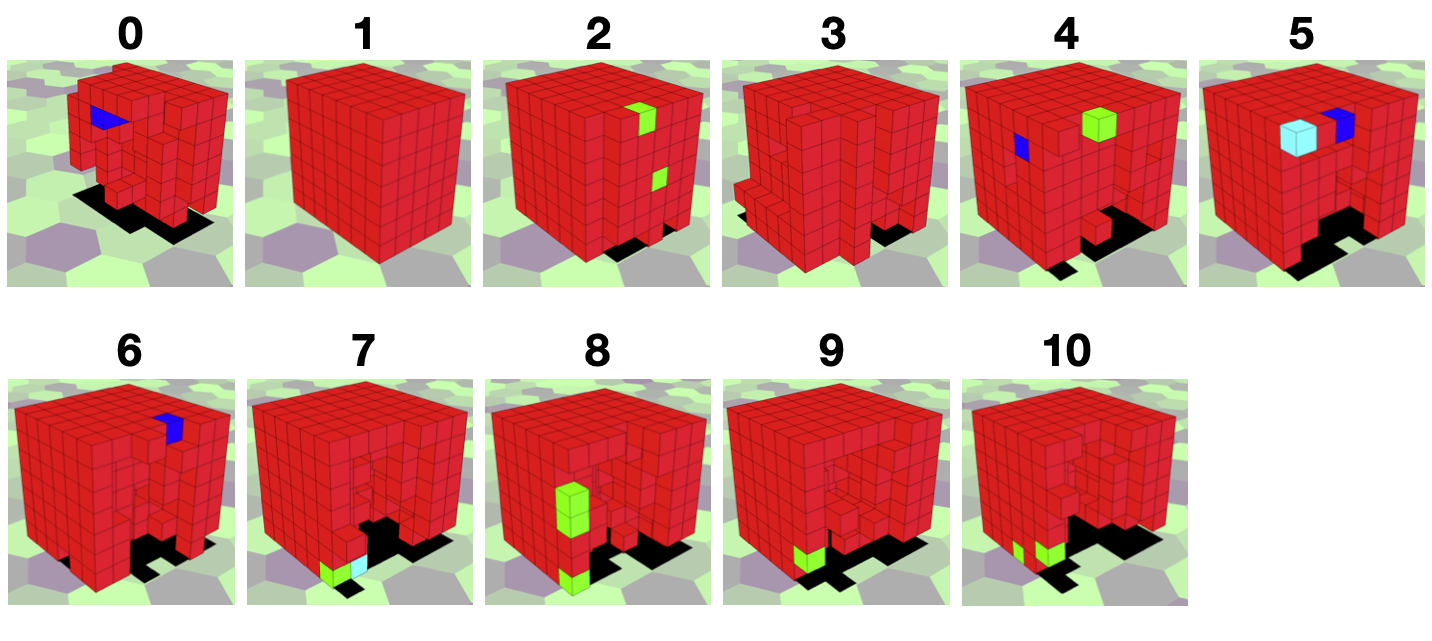}
        \caption{Multiped regereration}
        \label{Multiped regeneration}
    \end{subfigure}
    \newline
    \begin{subfigure}{1\textwidth}
        \centering
        \includegraphics[width=1\linewidth]{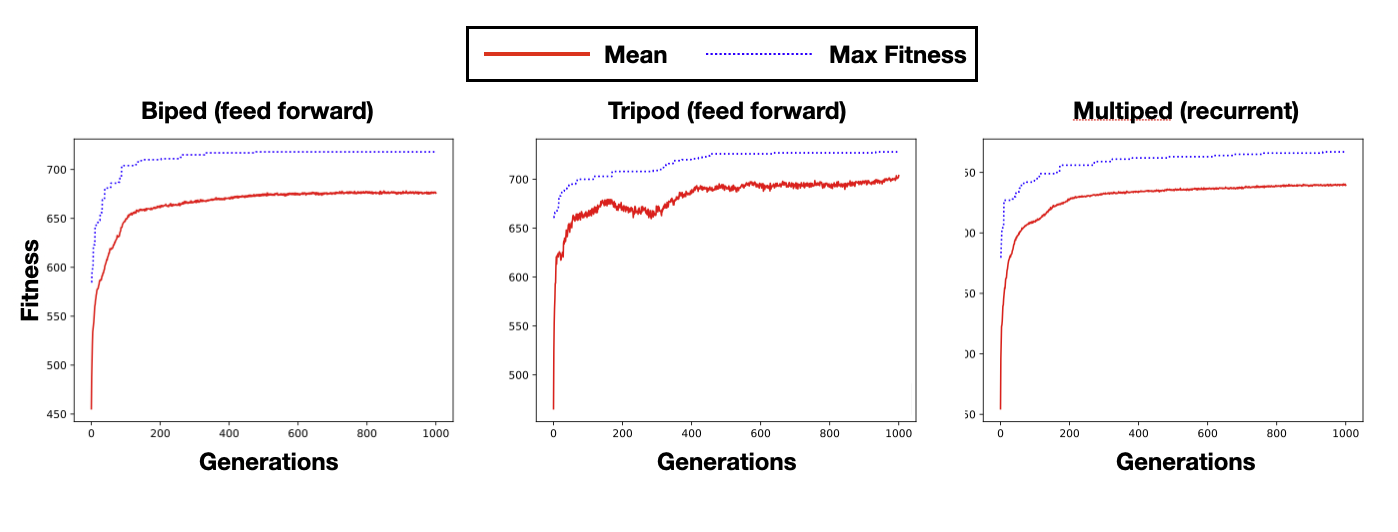}
        \caption{Fitness function of each morphology}
        \label{regeneration fitness}
    \end{subfigure}
    \caption{{\bf Regenerating soft robots} (a) Original, damaged, and regrown morphology. (b)-(d) Soft robot development after damage shown at different timesteps. (e) Training performance for recurrent/feed forward setup.}
    \label{regeneratio}
\end{figure}

% Set up for regenerating task
Here we investigate the ability of the soft robots to regenerate their body parts to recover from morphological damage. We chose three morphologies from the previous experiments, which are able to locomote well and as diverse as possible: the Biped (feed forward), Tripod (feed forward), and Multiped (recurrent). The morphologies of each of these three robots are shown in Fig.~\ref{regeneration set up} and the locomotion patterns in Fig.~\ref{3D Phylum}.

In these experiments, we damaged the morphologies such that one side of the robot was completely removed (Fig.~\ref{regeneration set up}). In the left side of these damaged morphologies, the cell states were set to empty and the maturation alpha values were set to zero. For the recurrent network, the memory of LSTM units in each cell were also reset to zero. 

We initially attempted regeneration using the original neural CAs of these three robots but regeneration failed and locomotion was not recovered. Therefore, we evolved another neural CA, which sole purpose it is to regrow a damaged morphology. In other words, one neural CA grows the initial morphology and the other CA is activated once the robot is damaged. 
%capable of growing an initial morphology and also displaying the ability to recover their morphology. 
Fitness for this second CA is determined by the voxel similarity between the original morphology and the recovered morphology (values in the range of $[0; 729]$). The maximum fitness of $ 9\times9\times9 = 729$ indicates that the regrown morphology is identical to the original morphology. We evolved these soft robots, which were allowed to grow for 10 steps, for $1,000$ generations with a population size of $1,000$. The next generation was selected from the top $20\%$.

\begin{table}[h]
    \centering
    \begin{tabular}{ |p{3.2cm}||p{2.5cm}|p{1.5cm}|p{1.5cm}|p{1.5cm}|}
    \hline
    \multicolumn{2}{|c|}{}& \multicolumn{3}{|c|}{Locomotion}\\
    \hline
    Morphology (Network) & Similarity & Original & Damaged & Regrown\\
    \hline
    Biped (feed forward) & $98\%$ (718/729) & 40.4  & 27.2($67\%$) & 35.1($86\%$) \\
    Tripod (feed forward) & $99\%$ (728/729) &44.5 & 1.63($3.6\%$) & 20.3($45\%$) \\
    Multiped (recurrent) &$91\%$ (667/729)& 42.7 & 5.36($12\%$) & 9.6($22\%$) \\
    % add similarity on table
    \hline
    \end{tabular}
    \caption{Morphology similarity and locomotion recovery rate.}
    \label{regeneration locomotion}
\end{table}

%Results similarity of morphology
For all three morphologies we trained both feed forward and recurrent neural CAs. The best performing network types for damage recovery were consistent with the original network type for locomotion in all morphologies (biped = feed forward, tripod = feedforward, multiped = recurrent ).
The results with the highest performing network type are summarised in Table~\ref{regeneration locomotion} and damaged morphologies for each of the robots are shown in Fig.~\ref{regeneration set up}. 
The results indicate that the Multiped was the hardest to reproduce, followed by the Biped and then the Tripod. The Tripod had a higher similarity than the other morphologies and the neural CA almost completely reproduced the original morphology with the exception of one cell. We hypothesise that regeneration for the Tripod is easier because it only requires the regrowth of one leg, a simple rod-like shape with only a few cells. 

\begin{figure}[htpb!]
    \centering
    \includegraphics[width=1\linewidth]{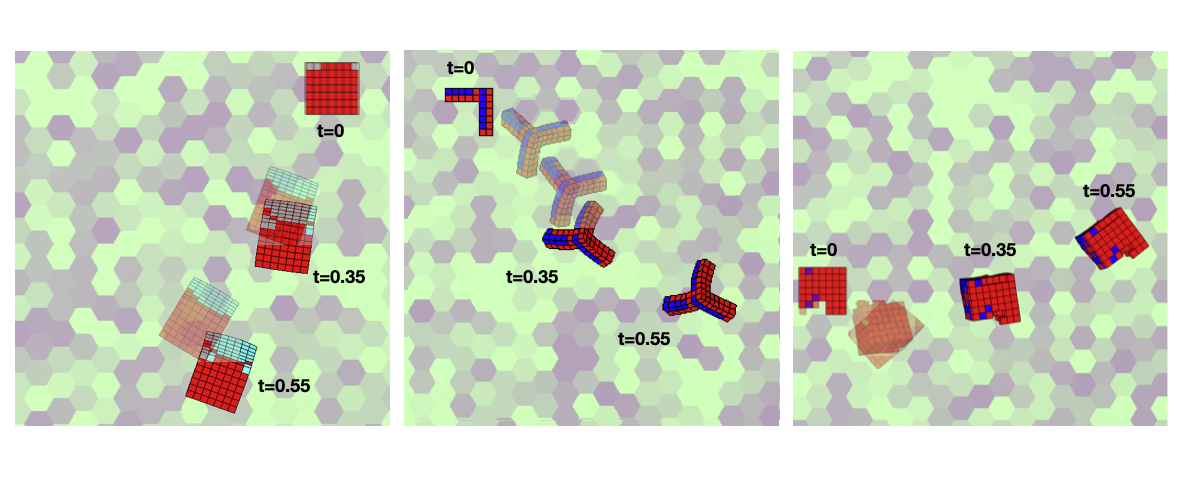}
    \caption{{\bf Recovery of locomotion. }The regrown morphologies are shown semi-translucent. From left to right: Biped, Tripod, and Multipod.}
    \label{recovery of locomotion}
\end{figure}

For comparison, we then measured the locomotion of the original, damaged, and regrown morphology with an evaluation time of $0.5$s for $10$ cycles in VoxCad. The ratio of regrowth and travel distance to the original morphology are shown in Table~\ref{regeneration locomotion} and its locomotion in Fig.~\ref{recovery of locomotion}. The damaged Biped maintained 67$\%$ of its original locomotion ability;  it replicated a similar locomotion pattern to the one observed in the L-Walker. As the Tripod lost one of its three legs, it was incapable of successful locomotion. Furthermore, the Multiped lost all locomotion -- the robot simply collapsed at the starting position. 

These results suggest that the location of the damage is important in determining how much the robot loses in terms of locomotion performance. For instance, in the case of the Biped, the left hand side and right hand side are symmetrical. This means that when the left hand side was removed, the right hand side was able to locomote in the same, almost unaffected way. Therefore, despite having the lowest similarity value  between the initial and regrown morphologies, there is little loss in performance. 
In contrast the Tripod regained less than half the locomotion of the original morphology, despite regaining its original morphology almost completely. It would appear that the one voxel it is unable to regenerate is necessary to prevent the robot from spinning, allowing it to move forward. The damage recovery results show  potential for soft robots capable of regrowth, but regrowth mechanisms that are not dependant of damage location are an important future research direction. 

\section{Discussion and Future Work}
The ability to control pattern formation is critical for the both the embryonic development of complex structures as well as for the regeneration/repair of damaged or missing tissues and organs. Inspired by this adaptive capacity of biological systems, in this paper we presented an approach for morphological regeneration applied to soft robots. 

We developed a new method for robot damage recovery based on neural cellular automata. %The approach is able to grow a variety of different soft robots  from a single cell, which was not possible with conventional soft robot simulations. 
While full regeneration is not always possible, the method shows promise in restoring the robot's locomotion ability after damage. The results indicate that the growth process can enhance the evolutionary potential of soft robots, and the regeneration of the soft robot's morphology and locomotion can provide some resilience to damage. 

The fitness landscape of the developmental evolving soft robot is likely very complex, and the simple evolutionary algorithm employed in this paper is therefore getting stuck in some of these local optima. This limitation could explain why we needed two neural CA, one for growing the initial morphology and one for regeneration, and why it was sometimes difficult for evolution to find a network that could completely replicate the original morphology for damage recovery.  We anticipate that the variety of quality diversity  approaches that reward more exploration during evolutionary search~\cite{Pugh2016}, such as  MAP-Elites~\cite{Mouret2015}, could allow for an even wider range of morphologies and escape some of these local optima.

Additionally, the locomotion and regeneration task in this paper is relatively simple.
%\todo{more types of damage. currently always cut the robot in half. Also drawback: two neural CAs are needed} 
Exciting future work will explore more complex tasks (e.g.\  recovery form more types of damage) that could benefit from morphological growth/regeneration, such as object manipulation, adaptation to environmental changes, task-based transformation, and self-replication.

Recently, soft robots designed using computer simulations have recently been recreated in real robot using a variety of materials~\cite{Howison2020}. With the development of material science, a variety of soft robots that can change their shape have been born~\cite{El-Atab2020}. Currently, the technology of tissue culture has been developed, and hybrid robots with dynamic plasticity are being developed~\cite{Kriegman2020c}.
In the future, it may be possible to create a hybrid robot that can grow spontaneously and recover its function from damage by creating a soft robot designed using the proposed model with living tissue. Because the approach presented in this paper only relies on the local communication of cells, it could be a promising approach for the next generation of these hybrid robots. 

\section*{Acknowledgements}
This work was supported by the Tobitate! (Leap for Tomorrow) Young Ambassador Program, a Sapere Aude: DFF-Starting Grant (9063-00046B), and KH's Academist supporters\footnote[1]{https://academist-cf.com/projects/119?lang=en}(Takaaki Aoki, Hirohito M. Kondo, Takeshi Oura, Yusuke Kajimoto,Ryuta Aoki). 
%Computation was provided by  ITU High Performance Cluster.

\bibliographystyle{splncs04}
\bibliography{main}
\end{document}